\documentclass[10pt,twocolumn,letterpaper]{article}

\usepackage[review,algorithms]{wacv}  
\usepackage[utf8]{inputenc}
\usepackage{textgreek}

\usepackage{graphicx}
\usepackage{amsmath}
\usepackage{amssymb}
\usepackage{booktabs}
\usepackage{pifont} % For symbols like \ding{108} (●)
\usepackage{textcomp}
\usepackage[table]{xcolor} % For HTML color model and \rowcolor
\definecolor{Gray}{gray}{0.9}
\usepackage{array} % For \arraybackslash in tabular columns
\usepackage{multirow} % For multirow cells in tables
\usepackage{etoolbox} % if not already loaded
\makeatletter
\@ifpackageloaded{lineno}{\AtBeginDocument{\nolinenumbers}}{}
\makeatother
\usepackage{xcolor}
\definecolor{setared}{HTML}{e74c3c}
\definecolor{setbblue}{HTML}{3498db}

\definecolor{wacvblue}{rgb}{0.21,0.49,0.74}
\usepackage[pagebackref,breaklinks,colorlinks,allcolors=wacvblue]{hyperref}

\date{} 

\title{DuPLUS: Dual-Prompt Vision-Language Framework for Universal Medical Image Segmentation and Prognosis}

\author{%
  {\small
  \textbf{Numan Saeed\textsuperscript{*} \quad %
  Tausifa Jan Saleem \quad %
  Fadillah Maani \quad %
  Muhammad Ridzuan \quad %
  Hu Wang \quad %
  Mohammad Yaqub}}\\[0.25em]
  {\small
  Department of Computer Vision, Mohamed bin Zayed University of Artificial Intelligence (MBZUAI), Abu Dhabi, UAE}\\
  {\footnotesize
  \textsuperscript{*}Corresponding author: \texttt{numan.saeed@mbzuai.ac.ae}}
}

\begin{document}
\maketitle
\begin{abstract}
 % Medical imaging is crucial in modern healthcare, yet deep learning models often remain task-specific, hindering generalizability and extensibility, and limiting their ability to provide diagnosis while neglecting crucial prognostic insights. Existing "universal" models for medical image analysis are limited by simplistic task conditioning and inadequate medical semantic understanding. 
 Deep learning for medical imaging is hampered by task-specific models that lack generalizability and prognostic capabilities, while existing 'universal' approaches suffer from simplistic conditioning and poor medical semantic understanding. To address these limitations, we introduce DuPLUS, a deep learning framework for efficient multimodal medical image analysis. DuPLUS introduces a novel vision-language framework that leverages hierarchical semantic prompts for fine-grained control over the analysis task, a capability absent in prior universal models. To enable extensibility to other medical tasks, it includes a hierarchical, text-controlled architecture driven by a unique dual-prompt mechanism. For segmentation, DuPLUS is able to generalize across three imaging modalities, ten different anatomically various medical datasets, encompassing more than 30 organs and tumor types. It outperforms the state-of-the-art task-specific and universal models on 8 out of 10 datasets. We demonstrate extensibility of its text-controlled architecture by seamless integration of electronic health record (EHR) data for prognosis prediction, and on a head and neck cancer dataset, DuPLUS achieved a Concordance Index (CI) of 0.69. Parameter-efficient fine-tuning enables rapid adaptation to new tasks and modalities from varying centers, establishing DuPLUS as a versatile and clinically relevant solution for medical image analysis. The code for this work is made available at: \url{https://anonymous.4open.science/r/DuPLUS-6C52}
\end{abstract}
    
\section{Introduction}
\label{sec:intro}

According to Harvard Health Publishing, it is estimated that over 80 million CT scans are performed annually in the United States \cite{harvard_radiation_risk_imaging}. This high volume underscores the central role of medical imaging in modern healthcare, supporting clinicians not only in screening and diagnosis but also in treatment planning and disease monitoring \cite{WHO2025}. While anatomical segmentation remains fundamental for diagnosis \cite{ma2024segment}, the field is increasingly shifting from purely descriptive diagnostic tasks toward prognosis—predicting disease outcomes—to support more informed and personalized clinical decisions.

This evolution demands medical image analysis systems that can (1) interpret diverse imaging modalities such as CT, MRI, PET, ultrasound, and X-ray; (2) generalize across anatomies and clinical conditions; (3) perform both segmentation and prognosis; and (4) integrate visual data with textual information, such as electronic health records (EHRs) and clinical notes. These challenges are compounded by the need to operate at varying anatomical scales—from organs to tumors and lesions—adding further complexity \cite{walsh2021imaging}.

Despite significant progress, most deep learning models for medical imaging remain narrowly focused on isolated, task-specific applications. While many models have achieved human-level or even superhuman performance on certain segmentation tasks \cite{liu2021review,leopold2019pixelbnn,novikov2018fully}, their clinical utility is often limited by three key challenges. First, these models typically exhibit poor generalizability, struggling to transfer knowledge across different imaging modalities, anatomical structures, or institutional settings \cite{guan2021domain}. Second, scaling task-specific models to real-world clinical environments is impractical, as maintaining a large number of independent models for segmentation and prognosis becomes computationally and logistically cumbersome \cite{selvan2023operating,selvan2022carbon}. Finally, existing models focus largely and exclusively on disease and anatomical segmentation, offering little to no prognostic insight—an increasingly critical demand in AI-assisted healthcare. This limitation is exacerbated by the scarcity of datasets annotated for both segmentation and prognosis \cite{schafer2024overcoming,lee2024foundation,wang2021annotation}, further hindering their extensibility to outcome-predictive tasks. These gaps point to a critical need for models that not only generalize across modalities and anatomies but also provide clinically actionable prognostic insights.

To overcome these challenges, we propose \textbf{DuPLUS}, a unified deep learning framework for multimodal medical image analysis that seamlessly integrates and extends a 3D segmentation model to prognosis (and potentially other tasks) within a single architecture. DuPLUS is built upon a hierarchical, text-controlled design that leverages a novel dual-prompt mechanism. The first prompt (T1) describes the imaging modality and anatomical region (e.g., "A CT of the abdomen") and modulates the encoder-decoder backbone using Feature-wise Linear Modulation (FiLM) \cite{perez2018film} layers, producing context-aware feature representations. The second prompt (T2) specifies the segmentation target (e.g., "A CT of the liver") and directly controls the model’s output head, allowing for fine-grained, on-demand segmentation of specific structures. This separation of anatomical hierarchy  enables robust feature representation alongside fine-grained task control—capabilities that are largely absent from existing universal models. 

Beyond segmentation, DuPLUS is designed to be extensible. We demonstrate its architecture can also integrate structured patient data, such as EHRs, to support prognosis prediction. Importantly, DuPLUS enables rapid adaptation to new imaging modalities or clinical tasks through parameter-efficient fine-tuning of the text encoder, without the need to retrain the full model \cite{hu2022lora}. By combining hierarchical prompting with efficient adaptation and clinical extensibility, DuPLUS offers a scalable and versatile solution for comprehensive medical image analysis.

\begin{figure*}[t]
\includegraphics[width=0.97\textwidth]{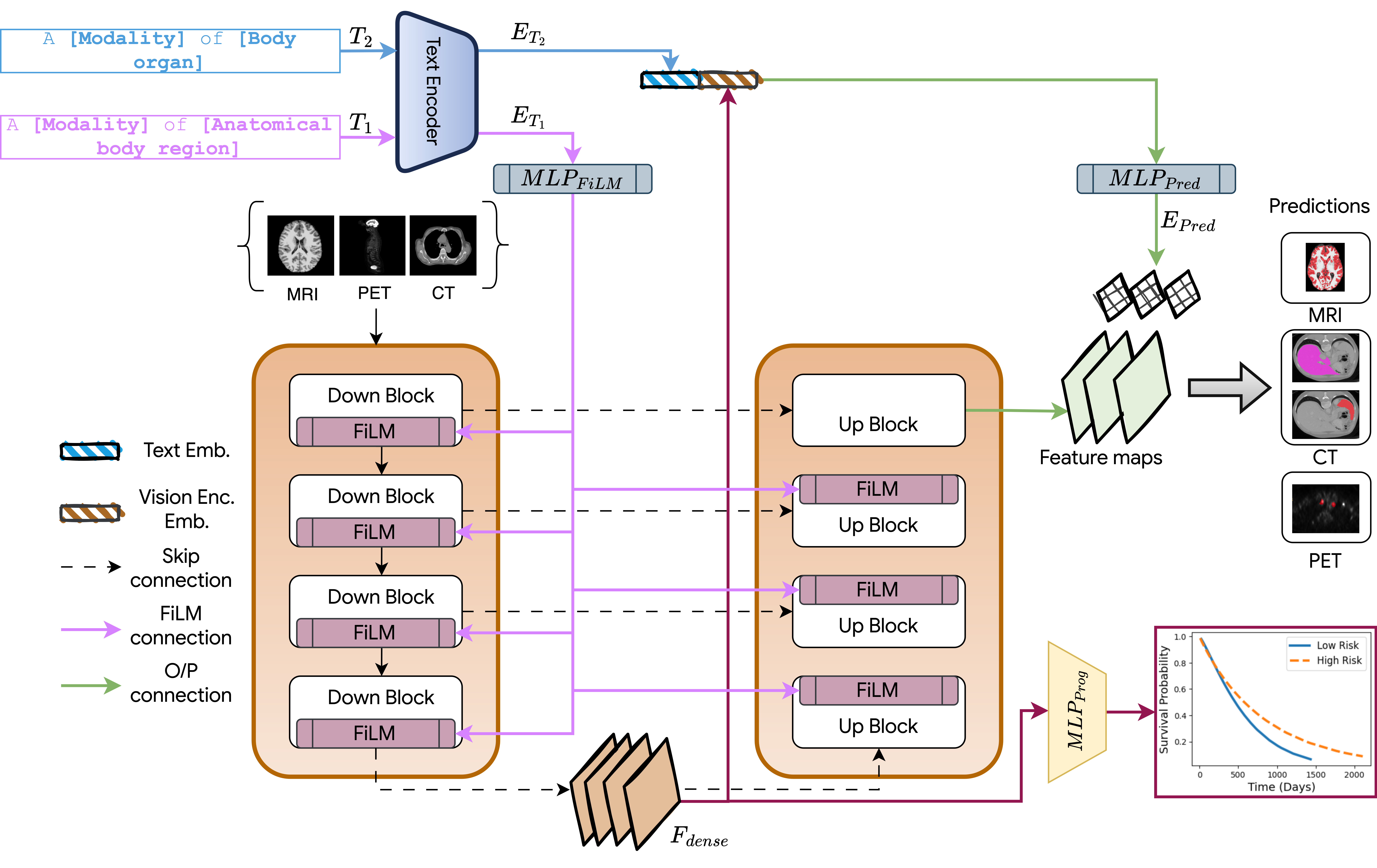}
\caption{Architecture of DuPLUS, a multimodal deep learning network controlled by text prompts. This diagram showcases DuPLUS's key components: the dual-prompt mechanism for text control and the FiLM layers for modality adaptation. It also illustrates the model's extensibility to prognosis prediction via a dedicated prediction module.} \label{main}
\end{figure*}

\section{Related Work}

\paragraph{Universal models.}  
Early attempts at universality in medical image segmentation began with DoDNet \cite{zhang2021dodnet}, which introduced one-hot task priors into a shared encoder–decoder to enable multi-organ and tumor segmentation under partially labeled abdominal CT. While effective, the orthogonality of one-hot encodings discarded potential relationships between tasks, and the approach remained confined to CT, limiting generalization across modalities and anatomies.  

Building on the need for broader representation learning, UniMiSS \cite{xie2022unimiss} proposed a Medical Transformer backbone with switchable patch embeddings that adaptively handle both 2D (e.g., chest X-rays) and 3D (e.g., CT volumes) inputs. By bridging the dimensionality gap, UniMiSS enabled cross-dimension self-distillation and leveraged abundant 2D data to compensate for the scarcity of 3D medical datasets. Pre-trained on over 100k 2D images and 5k 3D volumes, UniMiSS demonstrated positive gains across six downstream tasks, outperforming ImageNet pretraining and state-of-the-art self-supervised methods. However, its reliance on cross-dimension distillation restricted scalability when clinical labels were scarce, and it lacked explicit task- or modality-specific conditioning, limiting its ability to capture richer medical semantics.  

The CLIP-driven universal model \cite{liu2023clip} advanced conditioning by injecting semantic embeddings from a pretrained CLIP text encoder. By coupling image features with textual prompts, it captured anatomical relationships (e.g. liver, liver tumor, and hepatic vessel), jointly learned 25 organs and six tumor types across 14 CT datasets (more than 3,400 scans), and supported zero-shot prompting for unseen targets. Despite these advances, its reliance on a CLIP encoder trained on general-domain corpora hindered alignment with complex medical semantics, reducing its effectiveness in clinical settings.  

To address the need for improved generalizability across diverse modalities and tasks, the UniSeg model was introduced \cite{liu2023uniseg}. Its core mechanism is a universal prompt that encapsulates the correlations between different segmentation tasks. By combining this prompt with image features, UniSeg generates task-specific prompts to guide its decoder. Although effective, the application of the model has thus far been restricted to medical image segmentation.

Most recently, Hermes \cite{gao2024training} proposed context-prior learning inspired by radiology residency training. Hermes introduces a pool of learnable task and modality priors fused with image features via bidirectional attention, capturing complex inter-task and inter-modality relationships directly from medical data. Trained on 2,438 3D scans from 11 datasets spanning CT, PET, T1/T2, and cine MRI, Hermes achieved state-of-the-art segmentation while demonstrating scalability, transfer learning, and incremental learning. By learning priors directly from data, Hermes overcame the rigidity of one-hot conditioning and the domain gap of CLIP embeddings. Nevertheless, like its predecessors, Hermes remains predominantly segmentation-centric and does not explicitly integrate prognostic or multimodal clinical insights, limiting its extensibility to broader clinical decision-making.  
\paragraph{Transfer learning through multi-dataset pretraining.}  
Med3D \cite{chen2019med3d} addresses the lack of large 3D pretraining corpora by constructing 3DSeg-8 (eight CT/MRI datasets; brain, heart, liver, pancreas, spleen, vessels, prostate, hippocampus). A shared encoder with a multi-branch decoder (one branch per dataset) avoids label conflicts from incomplete annotations and yields transferable volumetric features. Med3D accelerates convergence up to an order of magnitude and improves accuracy (3–20\%) across lung segmentation, pulmonary nodule classification, and liver segmentation. On LiTS, a single-network Med3D+DenseASPP approaches the performance of the state-of-the-art ensemble. Building on dataset aggregation, MultiTalent \cite{ulrich2023multitalent} embraces contradictory protocols and overlapping classes (e.g., liver / liver vessel / liver tumor) by retaining dataset-specific semantics. It replaces softmax with sigmoid multi-head outputs to allow multiple labels per voxel and introduces a dataset- and class-adaptive loss for balanced supervision. Trained on 13 abdominal CT datasets (1,477 volumes; 47 classes), MultiTalent improves mean Dice over strong baselines, excels on difficult cancer categories, sets a new BTCV state-of-the-art, and serves as an efficient pretraining backbone. Yet, both Med3D and MultiTalent remain largely restricted to abdominal CT and fail to generalize across diverse modalities, limiting their universality.  
\paragraph{Self-configuring pipelines.}  
nnU-Net \cite{isensee2021nnu} establishes a self-configuring pipeline that maps a dataset “fingerprint’’ (spacing, size, class ratios) to a “pipeline fingerprint’’ (preprocessing, 2D/3D/cascade architectures, training, and post-processing). Decisions are split into fixed, rule-based, and empirical components. Up to three candidates are trained with cross-validation, and the best single/ensemble is selected, yielding near- or state-of-the-art performance across 23 public datasets (53 tasks) with standard compute. While a powerful automated baseline, its optimization is performed independently per dataset, without explicitly transferring knowledge across datasets. This dataset's isolation limits its potential for scaling to universal and extensible applications.

% \paragraph{Transformer-based segmentation.}  
% Transformer-based U-Nets extend conventional convolutional architectures by capturing long-range dependencies and richer global context. nnFormer \cite{zhou2023nnformer} interleaves convolution and transformers with local/global self-attention and skip attention, outperforming prior transformer hybrids (e.g., TransUNet, Swin-UNETR) on BraTS/MSD, Synapse, and ACDC, and complementing nnU-Net via ensembling. UNETR++ \cite{shaker2024unetr++} improves accuracy–efficiency trade-offs with Efficient Paired Attention (EPA). Paired spatial (linear-complexity) and channel attention share Q/K projections to reduce parameters while enhancing feature synergy. Interleaved EPA blocks in a hierarchical encoder–decoder deliver strong results across Synapse, BTCV, ACDC, and BraTS with more than 70\% reductions in parameters/FLOPs compared to strong baselines. Nevertheless, these designs remain narrowly focused on segmentation benchmarks and do not address broader clinical objectives such as prognosis or multimodal integration.  

\section{Methodology}

\subsection{Problem Statement}

Conventional image-based medical segmentation models $\mathcal{F}^{I}_{\Theta}: \mathcal{X}_{I} \rightarrow \mathcal{M}$ often exhibit limitations in generalizability, being typically optimized for specific imaging modalities and fixed anatomical targets. Here, $\mathcal{F}^{I}_{\Theta}$ denotes a standard segmentation architecture with parameters $\Theta$, $\mathcal{X}_{I}$ represents the input space of medical images for a specific modality, and $\mathcal{M}$ denotes the output space of binary segmentation masks for individual anatomical structures. To overcome these limitations and enhance task versatility, our primary goal is to introduce DuPLUS, a novel text-controlled framework $\widetilde{\mathcal{F}}^{\mathcal{T}^{(2)}}_{\Phi}: \mathcal{T}^{(2)} \times \mathcal{X}_{I} \rightarrow \mathcal{M}$, where $\mathcal{T}^{(2)}$ is a space of pairs of textual prompts providing hierarchical instructions i.e., ($T_1, T_2) \in \mathcal{T}^{(2)}$. We designed DuPLUS such that it achieves several key objectives: 
\begin{itemize}
\item \textbf{modality-adaptive segmentation} across diverse imaging modalities within $\mathcal{X}_{I}$, 
\item \textbf{cross-modal disentanglement} to minimize modality interference, 
\item  \textbf{parameter-efficient fine-tuning} for rapid adaptation to new tasks and datasets, and 
\item  \textbf{extensibility to new tasks beyond segmentation and prognosis (shown in this work)}, such as classification, by leveraging learned representations.
\end{itemize}

\subsection{Proposed Methodology: DuPLUS}

DuPLUS, visually depicted in Figure \ref{main}, incorporates an encoder-decoder architecture that is to U-Net \cite{ronneberger2015u}, however,  significantly enhanced by \textbf{hierarchical dual text prompt conditioning}.

\subsubsection{Hierarchical Conditioning}

\noindent DuPLUS text conditioning employs two distinct prompts, $T_1$ and $T_2$, enabling task versatility.

\begin{itemize}
    \item \textbf{Prompt $T_1$: Modality and Anatomical Context.} Exemplified by "A CT scan of the abdomen", $T_1$ defines the broad medical image context, specifying the \textit{imaging modality} and \textit{anatomical body region}. $T_1$ conditions the \textbf{Vision Encoder and Decoder} via Feature-wise Linear Modulation (FiLM) \cite{perez2018film}, enabling body region and modality-adaptive feature processing within the model architecture. For prognosis, $T_1$ could specify a risk score prediction task with EHR, as will be detailed in the Extensibility section.

    \item \textbf{Prompt $T_2$: Target-Specific Instruction.} Exemplified by "A CT scan of a liver", $T_2$ provides target-specific instructions. For segmentation, $T_2$ specifies the \textit{target body organ}, conditioning the \textbf{Prediction Head}, and thus, guiding organ-specific segmentation mask generation. 
\end{itemize}

\noindent This hierarchical approach allows $T_1$ to establish a modality-aware and region-specific feature space for segmentation or prognosis (objective i), while $T_2$ refines the target, directing the network towards the organ to segment, mimicking the way radiologist review medical scans.

\subsubsection{Multi-Modal Vision Encoder}

\noindent The Vision Encoder, a U-Net variant, handles multi-modal inputs $\mathcal{X}_{I}$. Dedicated input blocks extract modality-specific features for each modality $m \in \{CT, MRI, PET\}$. Modality-specific input blocks reduce interference by isolating early-stage feature extraction (objective ii), while FiLM layers adapt shared encoders to modality-specific contexts.

\subsubsection{Text Conditioning Modules} 
\noindent DuPLUS employs a \textbf{ClipMD Text Encoder} \cite{glassberg2023increasing} to embed prompts $T_1$ and $T_2$, into $E_{T_1}$ and $E_{T_2}$. Pretrained on biomedical corpora and frozen during training, ClipMD ensures stable embeddings and improved medical text understanding.

\begin{itemize}
    \item \textbf{FiLM Parameter Generation:}  Embedding $E_{T_1}$ is processed by an $MLP_{FiLM}$ to generate FiLM parameters $(\gamma_{down}^{(j)}, \beta_{down}^{(j)})$ and $(\gamma_{up}^{(k)}, \beta_{up}^{(k)})$ for each Down and Up Block via projections:
    \begin{equation}
    (\gamma_{down/up}^{(j)}, \beta_{down/up}^{(j)}) = MLP_{FiLM}^{(down/up, j)}(E_{T_1}).
    \end{equation}

    \item  \textbf{Prediction Head Parameter:} Embedding $E_{T_2}$ is concatenated with the most dense representation of the input image, $F_{dense}$, extracted by the deepest FiLM layer and processed by $MLP_{Pred}$ to generate embedding $E_{Pred} = MLP_{Pred}([E_{T_2}; F_{dense}])$, which parameterizes the Prediction Head.

\end{itemize}

\subsubsection{Feature-wise Linear Modulation (FiLM)}

\noindent  DuPLUS employs sequential \textbf{Down Blocks} and \textbf{Up Blocks} in its encoder and decoder paths, respectively. Each block contains multiple convolutional layers for up-sampling/down-sampling feature maps and a \textbf{FiLM} layer. The convolutional layers extract image features, $\hat{F}_{block}$, and FiLM receives text-based conditioning from Prompt $T_1$ (represented by $\gamma$ and $\beta$). The FiLM operation is based on dynamically modulating $\hat{F}_{block}$ based on Prompt $T_1$, enabling modality-adaptive processing throughout the network. The FiLM operation is defined as:
\begin{equation}
\text{FiLM}( \hat{F}_{block}, \gamma, \beta ) = \gamma \odot \hat{F}_{block} + \beta.
\end{equation}

\begin{figure*}[t]
\includegraphics[width=\textwidth]{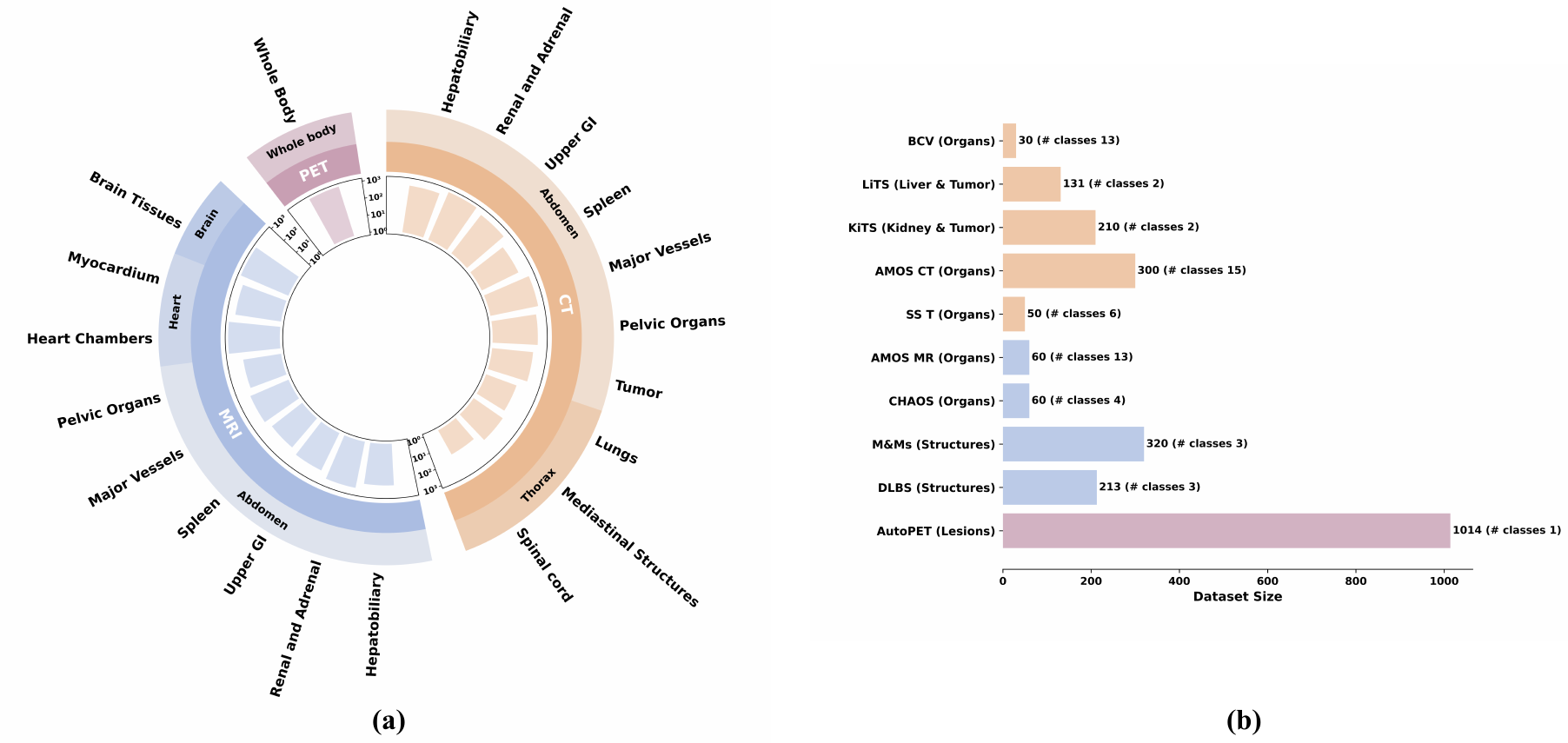}
\caption{A visualization of the used medical imaging datasets. (a) Distribution of anatomical classes across CT, MRI, and PET modalities. (b) Significant data imbalance in dataset sizes and classes of different structures is observed, which is a challenge for robust model training.} \label{dataset}
\end{figure*}

\subsubsection{Prediction Head for Segmentation}

\noindent For segmentation, the Prediction Head uses three $1 \times 1 \times 1$ convolutional layers, parameterized by embedding $E_{Pred}$. The prediction $P_k$ for organ class $k$ is:
\begin{equation}
P_k = \sigma( g(g(F * \theta_{k1}) * \theta_{k2}) * \theta_{k3} ),
\end{equation}
where kernel parameters $\theta_k = \{\theta_{k1}, \theta_{k2}, \theta_{k3}\}$ are derived from $E_{Pred}$, $g$ represents nonlinearity and $\sigma$ is the sigmoid function for the final binary segmentation mask. This design enables organ-specific adaptation without spatial distortion, balancing parameter efficiency and nonlinearity.

\subsubsection{Extensibility to Prognosis}

\noindent A key advantage of DuPLUS lies in its efficient adaptability to diverse clinical tasks, including patient outcome prediction (prognosis). Building upon a pre-trained DuPLUS for segmentation, we demonstrate that Low-Rank Adaptation (LoRA) \cite{hu2022lora} facilitates parameter-efficient fine-tuning of the text-understanding component of DuPLUS for prognosis prediction. Specifically, DuPLUS is presented with a structured textual prompt, such as: `\textit{Predict the risk score of a male patient, 52 years old, with CT imaging of the Head \& Neck, a weight of 82 kilograms, and a history of smoking and alcohol consumption}'. Subsequently, a dedicated prediction module ($MLP_{Prog}$) processes the image-derived features ($F_{dense}$, see Figure \ref{main}) to generate a quantitative risk score ($R_{risk}$):

\begin{equation}
R_{risk} = MLP_{Prog}(F_{dense}).
\end{equation}

% This demonstrates the \textit{continual learning} ability of DuPLUS – its capacity to learn new tasks efficiently without forgetting what it already knows.

\section{Experimental Setup}

\subsection{Datasets}
\noindent \textbf{Segmentation Datasets}: Our study utilized a diverse collection of ten publicly available datasets for segmentation, detailed in Figure \ref{dataset}. These datasets were selected to encompass a wide range of body regions (abdomen, thorax, head \& neck, brain, cardiac, whole body), imaging modalities (CT, MRI, cineMRI, PET, T1 \& T2 MRI), and clinical targets (organs, liver \& tumor, kidney \& tumor, lesions, structures), providing a robust testbed for evaluating the generalizability of DuPLUS. Additional details about these datasets are provided in the supplementary material (Section A). \noindent \textbf{Prognosis Dataset}: we use the HECKTOR dataset (Head \& neCK TumOR segmentation and outcome prediction) \cite{andrearczyk2021overview}. HECKTOR is a multimodal and multicenter head and neck cancer patient data set comprising co-registered CT and PET scans, along with corresponding segmentation masks (primary and nodule tumors) and EHR for 488 patients collected from seven centers.

\begin{table*}[!t]
    \centering
    \caption{Comparison of models across different datasets based on DSC (Dice Similarity Coefficient). All models were trained and evaluated under identical experimental conditions to ensure a fair comparison. Tradit. (traditional) refers to task specific models trained from scratch. }
    % \scriptsize % Reduce font size
    \renewcommand{\arraystretch}{1.2} % Increase row spacing for readability
    \definecolor{ctcolor}{HTML}{ebbb94}    % Peach color for CT
    \definecolor{mricolor}{HTML}{acbee3}   % Light blue for MRI
    \definecolor{petcolor}{HTML}{c9a0b4}   % Pink for PET
    \resizebox{\textwidth}{!}{%
    \begin{tabular}{>{\centering\arraybackslash}p{0.5cm}|>{\raggedright\arraybackslash}p{2.5cm}>{\columncolor{ctcolor}}c>{\columncolor{ctcolor}}c>{\columncolor{ctcolor}}c>{\columncolor{ctcolor}}c>{\columncolor{ctcolor}}c>{\columncolor{mricolor}}c>{\columncolor{mricolor}}c>{\columncolor{mricolor}}c>{\columncolor{petcolor}}c>{\columncolor{mricolor}}cc}
        \toprule
        \rowcolor{Gray}
        Cat & Models $\downarrow$ & BCV & SS T & LiTS T & KiTS T & AMOS CT & AMOS MR & CHAOS & M\&Ms & AutoPET & DLBS \\
        \midrule
        \multirow{2}{*}{\rotatebox[origin=c]{90}{Tradit.}} & nnUNet \cite{isensee2021nnu} & 84.23 & 88.53 & 64.91 & 81.72 & 88.79 & 85.49 & 91.34 & 85.65 & 65.43 & 94.22 \\
        & ResUNet & 84.36 & 88.59 & 64.87 & 81.89 & 88.97 & 85.43 & 91.34 & 85.73 & 65.52 & 94.31 \\
        \midrule
        \multirow{6}{*}{\rotatebox[origin=c]{90}{Baseline}} & Multi-decoder \cite{chen2019med3d} & 83.90 & 89.18 & 65.74 & 81.66 & 89.27 & 85.65 & 91.56 & 86.00 & 66.06 & 94.71  \\
        & DoDNet \cite{zhang2021dodnet} & 85.02 & 88.87 & 65.84 & 82.65 & 88.86 & 86.22 & 91.35 & 85.97 & 67.49 & 94.94  \\
        & CLIP-driven \cite{liu2023clip} & 85.12 & 89.34 & 65.37 & 82.83 & 88.94 & 86.39 & 91.81 & 86.04 & 66.78 & 95.17 \\
        & MultiTalent \cite{ulrich2023multitalent} & 85.18 & 89.18 & 65.33 & 82.25 & 89.13 & 86.57 & 91.55 & 86.28 & 71.51 & 95.75  \\
        & UniSeg \cite{liu2023uniseg} & 85.32 & 89.39 & 65.80 & 82.96 & 89.17 & 86.55 & 91.85 & 86.26 & 70.12 & 95.34  \\
        & Hermes-R \cite{gao2024training} & 85.99 & 89.50 & 67.49 & \textbf{85.53} & 89.63 & 86.78 & 92.01 & 86.94 & \textbf{73.69} & 96.21 \\
        \midrule
        \rotatebox[origin=c]{90}{Ours} & DuPLUS & \textbf{86.71} & \textbf{90.32} & \textbf{68.61} & 81.34 & \textbf{90.73} & \textbf{87.98} & \textbf{92.51} & \textbf{87.19} & 70.61 & \textbf{97.11}  \\
        \bottomrule
    \end{tabular}%
    }
    \label{tab:comparison}
\end{table*}

\begin{figure*}[!t]
\includegraphics[width=\textwidth]{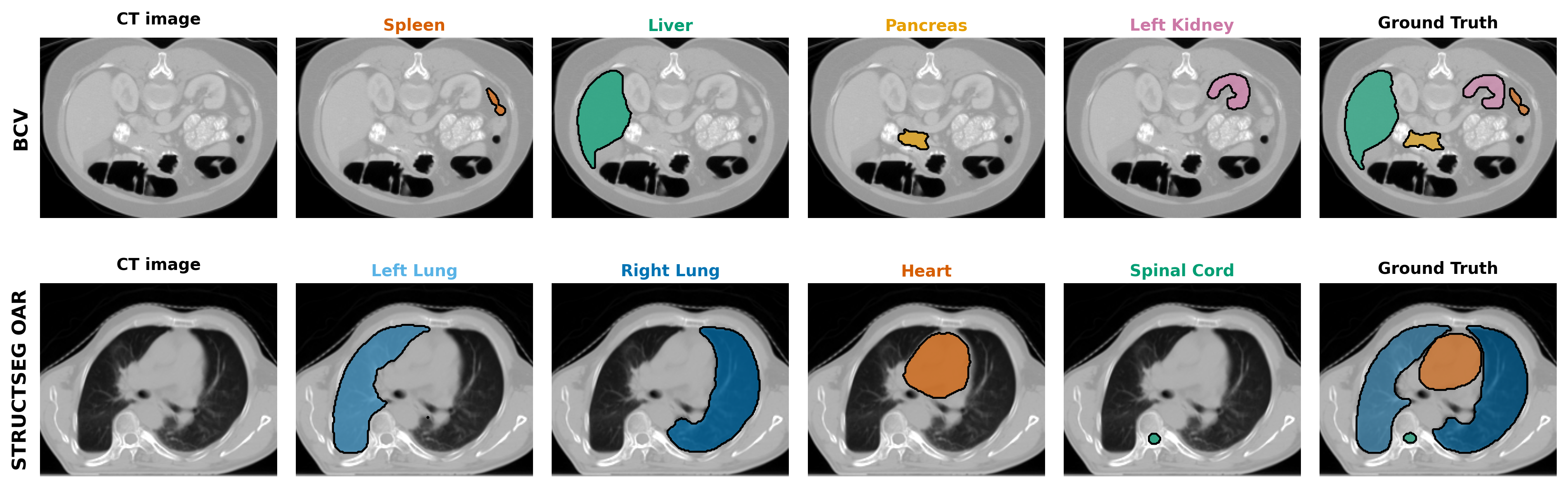}
\caption{Controllable, dual prompt-driven organ segmentation across datasets. Each row shows one CT dataset (BCV abdomen; STRUCTSEG OAR thorax). Columns: CT image, four prompted predictions, and Ground Truth. Inference uses two text prompts: (T1) a modality/region context, fixed per row (“A computed tomography of abdomen” for BCV; “A computed tomography of thorax” for STRUCTSEG OAR); and (T2) a target-organ prompt that is changed per column (“A computed tomography of spleen/liver/pancreas/left kidney” in BCV; “left lung/right lung/heart/spinal cord” in OAR). Holding T1 constant and switching only T2 deterministically switches the predicted structure on the same slice, demonstrating fine-grained text control without altering the image or model weights. Colored title fonts indicate the mask color for each organ; “CT image” and “Ground Truth” provide qualitative reference.}\label{qual1}
\end{figure*}

\begin{figure*}[!t]
\includegraphics[width=\textwidth]{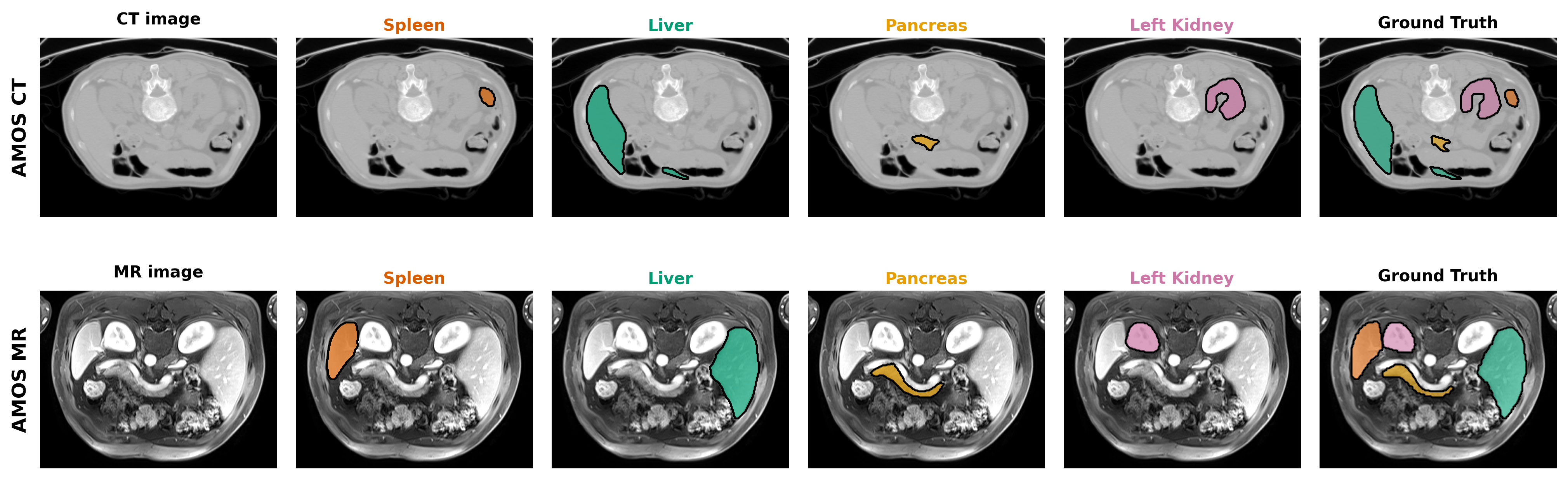}
\caption{Cross-modality organ segmentation using modality-aware text prompts. Rows show the same abdominal organs segmented from CT (top, AMOS CT) and MR (bottom, AMOS MR) images. The model uses dual prompts: T1 specifies the modality/region context, while T2 targets specific organs. By adapting T1 to the imaging modality ("computed tomography" vs "magnetic resonance"), the model successfully segments the same anatomical structures across both modalities without modality-specific training.}\label{qual2}
\end{figure*}

\subsection{Configurations, Implementation, and Baselines}

\noindent We implemented DuPLUS using PyTorch \cite{paszke2019pytorch} and conducted experiments on four NVIDIA RTX A6000 GPUs. All models, including DuPLUS and baselines (ResUNet, Multi-decoder \cite{chen2019med3d}, DoDNet \cite{zhang2021dodnet}, CLIP-driven \cite{liu2023clip}, UniSeg \cite{liu2023uniseg}, MultiTalent \cite{ulrich2023multitalent}, and Hermes-R \cite{gao2024training}), reimplemented with a ResUNet backbone for a fair comparison, were trained for 200 epochs, batch size of 16, a cosine learning rate scheduler (initial LR=2e-3), and on-the-fly data augmentations (random crop, rotation, scaling, brightness, contrast, gamma perturbation). Data preprocessing, applied uniformly across datasets to ensure a fair testbed, consisted of coordinate alignment, isotropic resampling to 1.5 × 1.5 × 1.5 mm, modality-specific intensity clipping (CT: [-990, 500], MR/PET: [2nd, 98th percentile]), and z-score normalization.  All models used a 128 × 128 × 128 input patch size and were evaluated using Dice Similarity Coefficient (DSC) on a 75\%/5\%/20\% training/validation/testing split \cite{gao2024training}. This uniform experimental framework ensured an unbiased comparison across network architectures, eliminating incentives related to patch size, spacing, augmentations, training, or evaluation procedures.

\subsection{Quantitative Evaluations}

\subsubsection{Universal Segmentation Performance}
Table~\ref{tab:comparison} presents a comparative analysis of DuPLUS against several state-of-the-art medical image segmentation models. The key strength of DuPLUS is not just its performance on a single task, but its consistent, state-of-the-art results across eight distinct datasets, spanning three imaging modalities (CT, MRI, PET) and over 30 different anatomical structures. This robust generalizability, powered by our dual-prompt mechanism, addresses a primary limitation of prior task-specific models.

The performance gains vary across datasets, but the model's consistency underscores its generalizability. DuPLUS showed its largest improvements on datasets with complex organ structures like DLBS (97.11\% DSC) and challenging tumor segmentation in LiTS (68.61\% DSC), suggesting our explicit text-conditioning excels at defining clear anatomical relationships. While DuPLUS demonstrated superior performance on 80\% of the tasks, the Hermes-R baseline was stronger on KITS T and AutoPET. We hypothesize that the implicit, learned context-priors of Hermes may be particularly effective for the subtle, varied boundaries found in tumors and PET lesions. However, the use of the ClipMD text encoder enables DuPLUS to effectively utilize semantic information from medical texts, enhancing its understanding of anatomical relationships and giving it an advantage over competitor models like Hermes-R, which do not accept text inputs.

\subsubsection{Extensibility and Fine-Tuning on HECKTOR}
To evaluate the adaptability of our framework, we extended the pre-trained model to new tasks on the HECKTOR dataset, which contains head and neck regions unseen during initial training. We assessed performance on both tumor segmentation and patient prognosis.

For segmentation, we compared DuPLUS against several established architectures that also utilize both CT and PET modalities. As shown in Table~\ref{tab:hecktor_seg}, our model outperforms these strong baselines. Our final result was achieved through a late fusion strategy. We first fine-tuned the DuPLUS model separately on CT and PET scans, which individually yielded average DSC scores of 58.4\% and 72.3\%, respectively. By ensembling the predictions from both modalities for each patient, we achieved our final score of 74\%. This approach, leveraging both structural detail from CT and functional activity from PET, demonstrates the effectiveness of our framework for multimodal data fusion.

\begin{table}[h!]
\centering
\caption{Head \& Neck Tumor Segmentation on HECKTOR.}
\label{tab:hecktor_seg}
\begin{tabular}{lc}
\hline
\textbf{Method} & \textbf{Dice Similarity (DSC)} \\ \hline
Swin UNETR \cite{hatamizadeh2021swin} & 0.70 \\
UNet \cite{meng2022radiomics} & 0.71 \\
3D ResUNet \cite{rezaeijo2022fusion} & 0.72 \\
\textbf{DuPLUS} & \textbf{0.74} \\ \hline
\end{tabular}
\end{table}

For patient prognosis, we integrated a DeepHit~\cite{lee2018deephit} prediction head and used CT, PET, and EHR data. As detailed in Table~\ref{tab:hecktor_prog}, DuPLUS surpasses common survival analysis baselines. This result is highly competitive and on par with the top-performing solutions on this dataset~\cite{andrearczyk2021overview}, confirming that our model's learned representations are effective for complex, multimodal prediction tasks beyond segmentation.

\begin{table}[h!]
\centering
\caption{Prognosis Performance on HECKTOR.}
\label{tab:hecktor_prog}
\begin{tabular}{lc}
\hline
\textbf{Method} & \textbf{Concordance Index (CI)} \\ \hline
MTLR \cite{fotso2018deep} & 0.63 \\
CoxPH \cite{cox1972regression} & 0.65 \\
DeepHit \cite{lee2018deephit} & 0.66 \\
\textbf{DuPLUS} & \textbf{0.69} \\ \hline
\end{tabular}
\end{table}

\begin{table*}[t]
    \centering
    \caption{Dual-prompt ablation across three datasets. Dice Similarity Coefficients (DSC) are reported in percentages (mean $\pm$ std). Parentheses give absolute ($\Delta$) changes from each dataset's baseline. For the ``Organ control'' condition, the notation shows the DSC on a present but unprompted organ versus the DSC on the prompted organ (e.g., L=Liver, S=Spleen, R=Right Lung).}
    \renewcommand{\arraystretch}{1.2} % Increase row spacing for readability
    \definecolor{ctcolor}{HTML}{ebbb94}   % Peach color for CT
    \definecolor{mricolor}{HTML}{acbee3}  % Light blue for MRI
    \resizebox{\textwidth}{!}{%
    \begin{tabular}{>{\raggedright\arraybackslash}p{4.5cm} >{\columncolor{ctcolor}}c >{\columncolor{ctcolor}}c >{\columncolor{mricolor}}c}
        \toprule
        \rowcolor{Gray}
        Condition & BCV (CT) & StructSeg Thorax (CT) & AMOS MR (MRI) \\
        \midrule
        Baseline & 86.71 $\pm$ 9.00 & 90.32 $\pm$ 7.00 & 87.98 $\pm$ 10.00 \\
        Modality T1 mismatch & 33.00 $\pm$ 30.00 (Δ $-$53.71) & 11.00 $\pm$ 13.00 (Δ $-$79.32) & 27.00 $\pm$ 25.00 (Δ $-$60.98) \\
        Modality T2 mismatch & 85.00 $\pm$ 10.00 (Δ $-$1.71) & 88.00 $\pm$ 6.00 (Δ $-$2.32) & 86.00 $\pm$ 10.00 (Δ $-$1.98) \\
        Modality both mismatch & 33.00 $\pm$ 30.00 (Δ $-$53.71) & 10.00 $\pm$ 13.00 (Δ $-$80.32) & 30.00 $\pm$ 25.00 (Δ $-$57.98) \\
        Region T1 mismatch & 9.00 $\pm$ 13.00 (Δ $-$77.71) & 3.00 $\pm$ 5.00 (Δ $-$87.32) & 4.00 $\pm$ 5.00 (Δ $-$83.98) \\
        Organ control & L: 0.00 / S: 95.00 & R: 0.00 / L: 96.00 & L: 0.00 / S: 97.00 \\
        \bottomrule
    \end{tabular}%
    }
    \label{tab:dual_prompt_ablation_two_decimals}
\end{table*}

\subsection{Qualitative Results}
The qualitative results in Figure~\ref{qual1} and Figure~\ref{qual2} visually validate the effectiveness and flexibility of our dual-prompt architecture. They specifically showcase the model's precise, on-demand target control and its robust adaptability across different imaging modalities and organs.

Figure~\ref{qual1} illustrates the fine-grained control offered by our target-specific prompt, $T_2$. For the same input CT slice, simply altering the target organ in the prompt - from ``spleen'' to ``liver'' to ``pancreas'', deterministically switches the output model to accurately segment only the requested structure. This result highlights the model's ability to disentangle and isolate specific anatomical targets based purely on textual instruction, without any change to the input image or model weights.

The versatility of the framework across modalities is demonstrated in Figure~\ref{qual2}. By only modifying the context prompt, $T_1$, to reflect the imaging modality (``computed tomography'' versus ``magnetic resonance''), DuPLUS successfully segments the same set of abdominal organs in both CT and MR images. This proves that the model effectively leverages the contextual information in $T_1$ to adapt its feature extraction process, confirming its ability to generalize to diverse clinical imaging data. Collectively, these visual examples provide strong qualitative support for our quantitative findings and validate the hierarchical, text-controlled design of DuPLUS.

\subsection{Discussion}

Our results demonstrate that DuPLUS not only establishes a new state-of-the-art for universal medical image segmentation but also seamlessly extends to complex, multimodal tasks like prognosis prediction. The framework's success is not incidental; it is a direct consequence of its hierarchical, dual-prompt architecture, the necessity of which was confirmed by our ablation studies.

The ablation study in Table~\ref{tab:dual_prompt_ablation_two_decimals} validates the critical role of each prompt. The context prompt ($T_1$) is essential for conditioning the network; mismatching the image modality or anatomical region causes a catastrophic drop in performance. Notably, a modality mismatch in the target prompt ($T_2$) results in only a minor decrease in accuracy. This contrast underscores the hierarchical nature of our design: $T_1$ is responsible for establishing the broad imaging context, while $T_2$ provides precise, on-demand control within that context. The ``Organ control'' experiment is definitive proof of the model's steerability; even when multiple valid organs were present in an image, the model selectively segmented only the prompted organ while completely ignoring the others.

\begin{figure}[!t]
\centering
\includegraphics[width=\columnwidth]{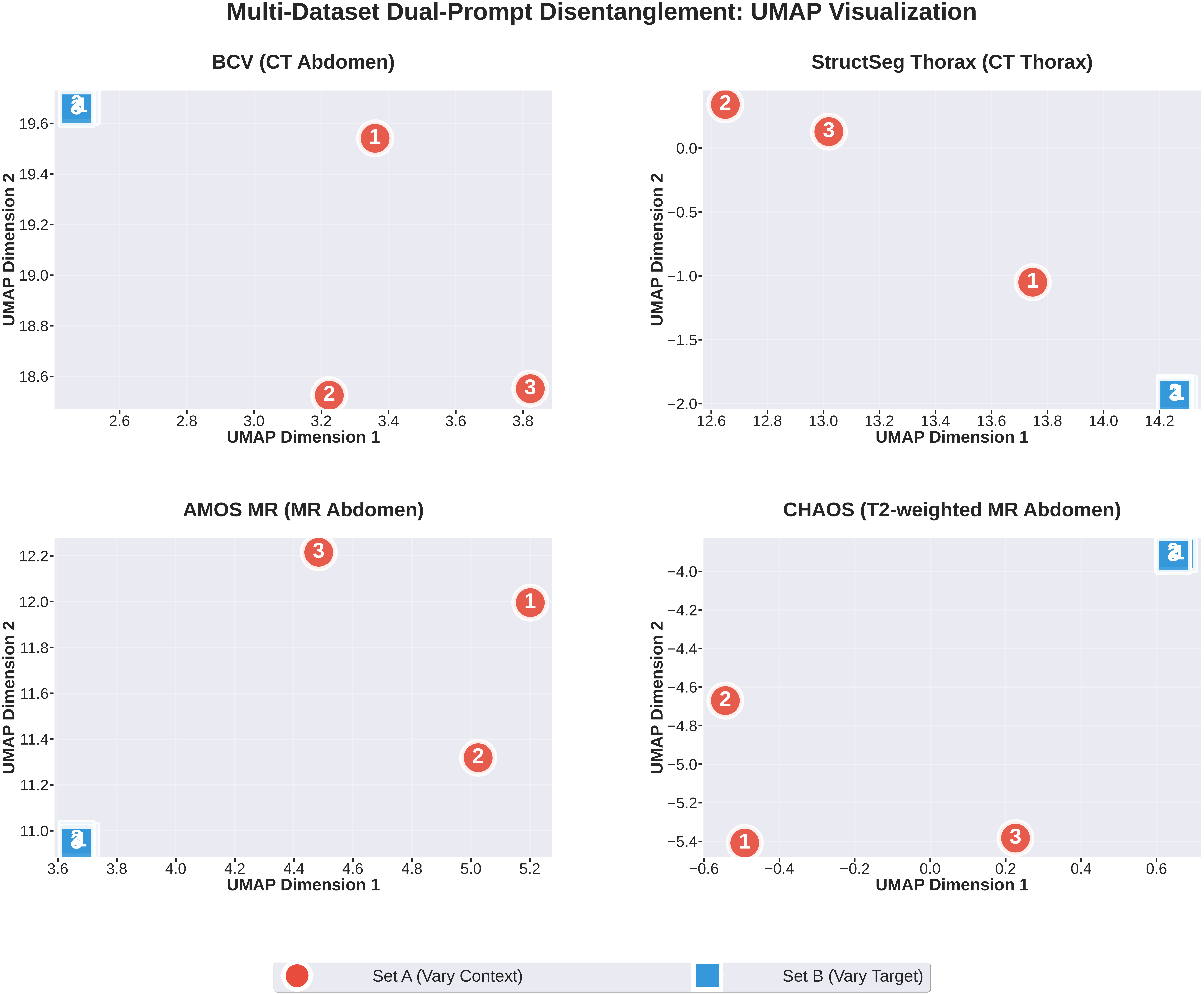}
\caption{\textbf{Dual-Prompt Feature Disentanglement Across Medical Imaging Modalities.} 
UMAP projection of bottleneck features under controlled prompt conditioning. 
\textcolor{setared}{\textbf{\Large$\bullet$}} \textbf{Set A:} Fixed target, varying context/modality prompts. 
\textcolor{setbblue}{\textbf{$\blacksquare$}} \textbf{Set B:} Fixed context, varying target prompts.} 
\label{fig:dual_prompt_disentanglement}
\end{figure}

Beyond demonstrating functional control, our analysis of the model's internal feature space in Figure~\ref{fig:dual_prompt_disentanglement} reveals \textit{how} this control is achieved. The UMAP visualizations show that DuPLUS learns to create disentangled feature representations. A fixed context ($T_1$) produces a stable and consistent feature cluster regardless of the target organ requested by $T_2$, suggesting the model separates the general representation of the image context from the specific segmentation task. This learned disentanglement is the core mechanism enabling DuPLUS’s robustness and generalizability across diverse tasks and modalities.

These findings have significant clinical implications. A steerable, multimodal model like DuPLUS paves the way for interactive diagnostic tools that allow clinicians to query medical images with text and receive integrated insights about both anatomy and patient outcomes. The model's extensibility to prognosis, achieved efficiently via Low-Rank Adaptation (LoRA) by updating only a fraction of the total parameters, further underscores its potential for creating scalable and sustainable clinical AI tools. A limitation of our current work is the reliance on structured text prompts. A compelling direction for future research is to enhance DuPLUS to interpret free-form text from clinical reports, further bridging the gap between AI models and real-world clinical workflows.

% Tight Set B clustering (similarity = 1.000) confirms successful disentanglement where identical contexts produce consistent representations regardless of target class. Distributed Set A points (similarity = 0.549-0.758) demonstrate contextual sensitivity while preserving target semantics. Validated across BCV (CT Abdomen), StructSeg (CT Thorax), AMOS (MR Abdomen), and CHAOS (T2 MR Abdomen) datasets.}

\section{Conclusion}
This work introduces DuPLUS, a novel hierarchical text-controlled framework for versatile medical image analysis. Comprehensive experimental results demonstrate that DuPLUS achieves state-of-the-art segmentation performance across diverse datasets and shows strong capabilities on multimodal data and prognosis prediction on the HECKTOR dataset. Future research may explore the potential of DuPLUS by evaluating its performance on classification tasks and integrating it into end-to-end clinical decision support systems. Furthermore, the proposed framework could be applied to a broader range of medical imaging datasets and clinical applications to fully assess its generalizability and clinical impact.
{
    \small
    
 % <- uses the .bbl you generated
}
\clearpage
\appendix
\section{Dataset details}
\label{sec:dataset}
This section provides a comprehensive overview of the segmentation datasets utilized in our study, with details on subject numbers, imaging modalities, annotation schemes, and data sources.\par
\textbf{BCV}:  
The BCV dataset \cite{landman2015multi} comprises abdominal CT scans from 50 subjects, with 30 cases publicly released for training purposes. A total of thirteen abdominal organs were delineated in 3D using MIPAV software, covering the spleen, both kidneys, gallbladder, esophagus, liver, stomach, aorta, inferior vena cava, portal and splenic veins, pancreas, and both adrenal glands. In instances where an organ was absent (e.g., gallbladder or right kidney), it was excluded from labeling. The scans were originally acquired during routine clinical practice at Vanderbilt University Medical Center.

\textbf{LiTS}:
The LiTS dataset \cite{bilic2023liver} contains 201 abdominal CT volumes, of which 131 are provided for training and 70 for testing; annotations are available exclusively for the training set. Each case includes coarse liver segmentations along with fine-grained tumor masks. Data were collected across several international institutions, including Ludwig Maximilian University of Munich, Radboud University Medical Center Nijmegen, Polytechnique and CHUM Research Center Montréal, Tel Aviv University, Sheba Medical Center, IRCAD Institute Strasbourg, and the Hebrew University of Jerusalem. The cohort comprises patients diagnosed with liver tumors such as hepatocellular carcinoma (HCC) as well as secondary liver malignancies and metastatic disease originating from colorectal, breast, or lung primaries. The tumors display heterogeneous enhancement characteristics, encompassing both hyperdense and hypodense appearances. The collection integrates pre- and post-treatment abdominal CT scans acquired using diverse scanners and imaging protocols. 

\textbf{KiTS19}: The KiTS19 dataset \cite{heller2019kits19} consists of segmented CT scans and corresponding treatment information from 300 patients who underwent either partial or radical nephrectomy for renal tumors at the University of Minnesota Medical Center between 2010 and 2018. Of these, 210 cases were made publicly accessible, while the remaining 90 were retained for evaluation. 

\textbf{AMOS CT}:
The AMOS CT \cite{ji2022amos} subset contains 500 abdominal CT scans collected from patients with tumors or other abnormalities at Longgang District People’s Hospital, using eight different scanners and vendors. Each case includes annotations for 15 organs: spleen, right and left kidneys, gallbladder, esophagus, liver, stomach, aorta, inferior vena cava, pancreas, right and left adrenal glands, duodenum, bladder, and prostate/uterus. 

\textbf{AMOS MR }:
The AMOS MR \cite{ji2022amos} subset comprises 100 abdominal MRI scans acquired from the same clinical source and scanner diversity as the AMOS CT subset. Manual annotations are provided for 15 organs, but some cases in the validation set lack bladder and prostate labels, restricting MRI segmentation to 13 organ categories. 

\textbf{StructSeg (SS T)}:  
The SS T subset of the StructSeg dataset \cite{li2020automatic} originated from the StructSeg challenge on organ-at-risk (OAR) and gross target volume (GTV) segmentation for radiation therapy planning in lung and nasopharynx cancers. Specifically, SS T focuses on thoracic OAR segmentation using CT scans from 50 lung cancer patients. Each scan is manually annotated for six critical OARs: left lung, right lung, spinal cord, esophagus, heart, and trachea.  

\textbf{CHAOS}:
The CHAOS dataset \cite{kavur2021chaos} originated from a challenge aimed at abdominal organ segmentation. For this study, we utilize Task 5, which includes MRI scans of 20 subjects acquired in three sequences: T1-in-phase, T1-out-phase, and T2-SPIR. Annotations are provided for four abdominal organs including liver, spleen, and both kidneys.  

\textbf{M\&Ms}:
The M\&Ms dataset \cite{campello2021multi} was developed for the MICCAI 2020 challenge on cardiac magnetic resonance (CMR) segmentation. The dataset includes both healthy individuals and patients diagnosed with hypertrophic and dilated cardiomyopathy. Data acquisition took place across clinical sites in Spain, Germany, and Canada, using scanners from four vendors: Siemens, GE, Philips, and Canon. The training portion consists of 150 annotated studies, while the remaining 170 cases are reserved for testing. Manual annotations include three cardiac structures, left ventricle, right ventricle, and left ventricular myocardium at both end-diastolic and end-systolic phases.

\textbf{DLBS}:  
The Dallas Lifespan Brain Study (DLBS) \cite{rodrigue2012beta} is a longitudinal neuroimaging project designed to investigate the preservation and decline of cognitive function across the adult lifespan. A central focus of the study is on resilience mechanisms and the early trajectories that may lead toward Alzheimer’s disease. For our work, we utilize 213 T1-weighted MRI scans from the DLBS cohort, which include manual segmentations of cerebrospinal fluid, gray matter, and white matter.

\textbf{AutoPET}:  
The AutoPET dataset \cite{gatidis2022whole} provides 1,014 annotated whole-body Fluorodeoxyglucose (FDG) PET/CT studies. Among these, 501 scans are from patients diagnosed with malignant lymphoma, melanoma, or non-small cell lung cancer (NSCLC), while the remaining 513 serve as negative control studies without PET-positive malignant lesions.

\end{document}